\documentclass[runningheads]{llncs}
\usepackage[T1]{fontenc}
\usepackage{graphicx,verbatim}
\usepackage{enumitem}
\usepackage{amssymb}
\usepackage{amsmath}
\usepackage{booktabs}
\usepackage{graphicx}
\usepackage{bbding}
\usepackage{utfsym}
\usepackage{hyperref}
\usepackage{url}
\begin{document}
\title{XFMamba: Cross-Fusion Mamba for Multi-View Medical Image Classification}
%

\author{
    Xiaoyu Zheng\inst{1} \and 
    Xu Chen\inst{1,4} \and 
    Shaogang Gong\inst{2} \and 
    Xavier Griffin\inst{3} \and 
    Greg Slabaugh\inst{1}
}

\authorrunning{Zheng et al.}  

\institute{
    Digital Environment Research Institute (DERI), Queen Mary University of London, London, UK \and
    School of Electronic Engineering and Computer Science, Queen Mary University of London, London, UK \and
    Blizard Institute - Faculty of Medicine and Dentistry, Queen Mary University of London, London, UK \and
    Department of Medicine, University of Cambridge, Cambridge, UK \\
    \email{xiaoyu.zheng@qmul.ac.uk}  
}

\maketitle              
\begin{abstract}
Compared to single view medical image classification, using multiple views can significantly enhance predictive accuracy as it can account for the complementarity of each view while leveraging correlations between views. Existing multi-view approaches typically employ separate convolutional or transformer branches combined with simplistic feature fusion strategies. However, these approaches inadvertently disregard essential cross-view correlations, leading to suboptimal classification performance, and suffer from challenges with limited receptive field (CNNs) or quadratic computational complexity (transformers). Inspired by state space sequence models, we propose \emph{XFMamba}, a pure Mamba-based cross-fusion architecture to address the challenge of multi-view medical image classification. XFMamba introduces a novel two-stage fusion strategy, facilitating the learning of single-view features and their cross-view disparity. This mechanism captures spatially long-range dependencies in each view while enhancing seamless information transfer between views. Results on three public datasets, MURA, CheXpert and DDSM, illustrate the effectiveness of our approach across diverse multi-view medical image classification tasks, showing that it outperforms existing convolution-based and transformer-based multi-view methods. Code is available at \url{https://github.com/XZheng0427/XFMamba}.

\keywords{Multi-view \and Cross fusion \and Medical image classification \and State space models.}
\end{abstract}
\section{Introduction}
In many clinical applications, such as the detection of orthopedic fractures, radiologists analyze multiple views within a diagnostic examination. Typically each view provides complementary insights, but their correlation can also be leveraged to improve predictive accuracy. 
Numerous recent studies aim to develop multi-view networks inspired by the multi-view analysis performed by radiologists, leading to the creation of more robust and performant models~\cite{1,2}. 

Previous approaches in multi-view image analysis rely on Convolutional Neural Networks (CNN) or Transformers. While CNN-based approaches are recognized for their scalability and linear computational complexity in the number of image pixels, their reliance on local receptive fields introduces a bias toward local feature extraction, limiting their ability to directly capture long-range dependencies~\cite{3,4}. Moreover, CNNs often employ a weight-sharing mechanism across different view features of the input image, limiting their flexibility in adapting to unseen or low-quality medical images. To address the limited context understanding of CNNs, methods such as CVT~\cite{5} and MV-HFMD~\cite{25} utilize CNNs as the backbone for local feature extraction while incorporating cross-view attention mechanisms to transfer features between unregistered views.  With the emergence of Transformers, transformer-based methods such as T-MVF~\cite{7} enhance visual modeling capabilities by leveraging global view features and dynamically adaptable weights. However, the attention mechanism in transformer-based methods results in quadratic computational complexity relative to input size and limitations on token size, posing substantial challenges to efficiency~\cite{8}. Building on the architecture of vision transformers, multi-view Swin Transformer approaches such as MV-Swin-T~\cite{6} and WT-MVSNet~\cite{9} aim to improve efficiency by reducing the dimensions or strides of processing windows. However, this optimization comes at the cost of limiting cross-view feature interactions.

To address the aforementioned limitations, Selective Structured State Space Models (S6), also referred to as Mamba~\cite{8}, have garnered increasing attention for their ability to selectively forget or propagate information, achieve global feature coverage, and utilize dynamic weights with linear computational complexity. Mamba has demonstrated exceptional performance in natural language processing~\cite{8} and computer vision tasks~\cite{11,17}. Furthermore, extensive research has investigated its potential in medical imaging, including image segmentation~\cite{10} and image classification~\cite{12}. Multi-view BI-Mamba~\cite{26} employs an early fusion mechanism to capture and integrate multi-view information. However, these recent studies incorporate Mamba as a plug-and-play module without a task-specific, in-depth design for multi-view medical imaging. Additionally, the exploration of Mamba in multi-view medical imaging tasks for cross-image fusion remains limited, particularly in architectural adaptations and fusion strategies.

Inspired by these advantages, we propose XFMamba, a pure Mamba network for multi-view cross-fusion, tailored to address the challenges of multi-view unregistered medical image classification. XFMamba integrates a four-stage encoder and two-stage fusion modules. The encoder effectively captures multi-scale features across multiple views. Furthermore, the two-stage fusion module is designed to exchange and align cross-view information and enhance multi-view integration.
Comprehensive experiments conducted on the MURA musculoskeletal radiographs dataset~\cite{13}, the CheXpert chest X-ray dataset~\cite{14}, and the CBIS-DDSM mammography dataset~\cite{15} demonstrate that XFMamba outperforms state-of-the-art models across three varied clinical problems. The key contributions of this work can be summarized as follows: 
\begin{enumerate}
    \item Our cross-fusion work marks the successful application of state space models, specifically Mamba, in multi-view medical image classification.
    \item We propose a two-stage, Mamba-based fusion mechanism designed to efficiently extract and seamlessly integrate information across multiple views. 
    \item Comprehensive evaluations across three varied datasets demonstrate the superior accuracy and efficiency of our method, establishing a new benchmark for Mamba's potential in the multi-view medical imaging domain.
\end{enumerate}

\section{Methods}
\subsection{Preliminaries}
\subsubsection{State Space Models} State Space Models (SSMs)~\cite{16} constitute a category of sequence-to-sequence modeling frameworks distinguished by their constant dynamics over time, a characteristic commonly referred to as linear time-invariance. SSMs, with their linear computational complexity, efficiently capture underlying dynamics by implicitly mapping to latent states, formally expressed as:
\begin{equation}
\label{eq1}
h'(t)=\textbf{A}h(t)+\textbf{B}x(t), y(t)=\textbf{C}h(t)+\textbf{D}x(t).
\end{equation}
Where, $x(t)\in \mathbb{R}$ denotes the input, $h(t)\in \mathbb{R}^N$ denotes the hidden state, and $y(t)\in \mathbb{R}$ denotes the output. $h'(t)$ refers to the time derivative of $h(t)$. Additionally, $\textbf{A}\in \mathbb{R}^{N \times N}$ refers to the state matrix. $\textbf{B}\in \mathbb{R}^{N \times 1}$, $\textbf{C}\in \mathbb{R}^{1 \times N}$ and $\textbf{D}\in \mathbb{R}$ denote the projection matrices.

Mamba~\cite{8} employs the zero-order hold (ZOH) discretization method to convert ordinary differential equations (ODEs) into discrete functions, making it particularly well-suited for deep learning applications. The discretization process applies a time step $\Delta$ to transform the continuous state parameters $\textbf{A}$ and $\textbf{B}$ into discrete parameters $\overline{\textbf{A}}$ and $\overline{\textbf{B}}$, which can be defined as:
\begin{equation}
\overline{\textbf{A}}=exp(\Delta \textbf{A}), \overline{\textbf{B}}=(\Delta \textbf{A})^{-1}exp(\Delta \textbf{A}-I)\cdot \Delta\textbf{B}. 
\end{equation}
Then, Eq.~(\ref{eq1}) takes the following form:
\begin{equation}
\label{eq3}
h_{t}=\overline{\textbf{A}}h_{t-1}+\overline{\textbf{B}}x_{t}, y_{t}=\textbf{C}h_{t}+\textbf{D}x_{t}
\end{equation}
Eq.~(\ref{eq3}) represents the fundamental operation within the SSM module.

\subsection{Proposed Architecture}
\begin{figure}[!ht]
\includegraphics[width=\textwidth]{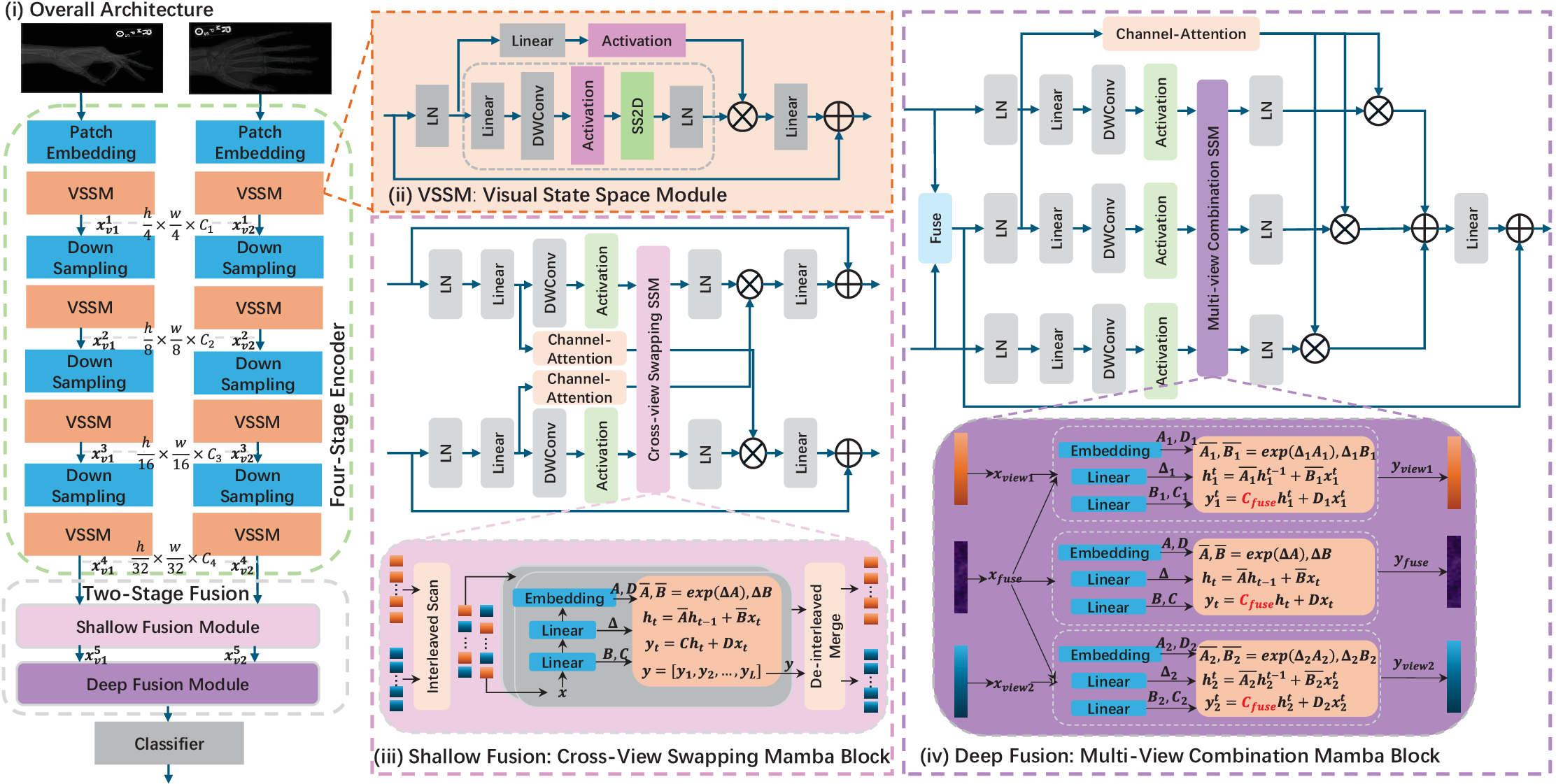}
\caption{XFMamba architecture. (i) The overall architecture is composed of a four-stage encoder and two-stage fusion module. (ii) Visual State Space Module (VSSM) for feature extraction. (iii) Cross-view swapping Mamba (CVSM) block for shallow fusion. (iv) Multi-view combination Mamba (MVCM) block for deep fusion.} 
\label{fig1}
\end{figure}

As shown in Fig.~\ref{fig1}, the proposed approach consists of a four-stage multi-scale encoder and a two-stage cascade fusion module. For each input image, the encoder is comprised of four visual state space modules (VSSMs), incorporating downsampling operations and sequentially cascaded to extract hierarchical image features across multiple levels. Next, features extracted from the two branches are processed and integrated through a two-stage cascade fusion module, which consists of a Cross-View Swapping Mamba (CVSM) block for local feature cross-view correlation (shallow fusion) and a Multi-View Combination Mamba (MVCM) block for holistic feature multi-view correlation (deep fusion). Finally, the refined and merged features are fed into a classifier to produce the prediction.

\subsubsection{XFMamba Encoder.} The multi-scale encoder consists of two four-stage feature extraction branches that process two greyscale images with different views as input, defined as $X_{v1},X_{v2}\in \mathbb{R}^{H \times W \times 1}$. The multi-scale encoder adopts VMamba as the backbone~\cite{17}. Inspired by ViT~\cite{18}, it starts by partitioning the input into patches to produce feature maps, i.e., $X^{1}_{v1,v2}\in \mathbb{R}^{\frac{H}{4} \times \frac{W}{4} \times C_1}$. Next, the features are progressively processed through three stages of downsampling and VSSM, as shown in Fig.~\ref{fig1}(i) and (ii), resulting in multi-scale feature representations, i.e., $X^{2}_{v1,v2}\in \mathbb{R}^{\frac{H}{8} \times \frac{W}{8} \times C_2}$, $X^{3}_{v1,v2}\in \mathbb{R}^{\frac{H}{16} \times \frac{W}{16} \times C_3}$, and $X^{4}_{v1,v2}\in \mathbb{R}^{\frac{H}{32} \times \frac{W}{32} \times C_4}$. We implement the VSSM using Selective Scan 2D (SS2D) modules proposed by VMamba~\cite{17}. As illustrated in Fig.~\ref{fig1}(ii), the features are processed through a sequence comprising linear projection (Linear), depth-wise convolution (DWConv), and the Sigmoid Linear Unit (SiLU) activation function. Subsequently, an SS2D module with a multiplicative connection is utilized to capture spatially long-range (wider context) information. 

\subsubsection{Shallow Fusion: Cross-View Swapping Mamba.} As illustrated in Fig.~\ref{fig1}(iii), we introduce a CVSM block that leverages an interleaving mechanism to enhance cross-view features by incorporating information from the other view. The two input features are initially processed through linear layers and depth-wise convolutions, followed by SiLU activation, and are then passed into the cross-view interleaving Selective Scan Module (SSM). Specifically, the cross-view interleaving SSM mechanism is firstly applied on $X^{4}_{v1}$ and $X^{4}_{v2}$, facilitating efficient feature interleaving across multiple views without incurring additional computational overhead, as shown in Eq.~\ref{eq4}.
\begin{equation}
\begingroup
\small   
\label{eq4}
\begin{aligned}
M(C) =
\begin{cases}
1, \text{if } C\ \text{mod } 2 = 0,\\
0, \text{if } C\ \text{mod } 2 = 1.
\end{cases} 
\widetilde{X}^{5}_{v1/v2}[:, C] =
\begin{cases}
\widetilde{X}^{4}_{v2/v1}[:, C],  \text{if } M(C) = 1, \\
\widetilde{X}^{4}_{v1/v2}[:, C],  \text{if } M(C) = 0. 
\end{cases}
\end{aligned}
\endgroup
\end{equation}
Here, $\widetilde{X}^{5}_{v1/v2},\widetilde{X}^{4}_{v1/v2} \in \mathbb{R}^{B \times N \times C}(N=H \times W)$ represent the flattened features. $M(C)$ defines the channel-wise mask, comprising binary values where $0$ signifies no interleaving and $1$ denotes interleaving. Subsequently, the interleaved cross-view features are processed through their corresponding Mamba blocks to extract long-range cross-view spatial information. The resulting sequences are then reshaped and merged to generate the output scanned features using the de-interleaving merge mechanism. This results in the features $X^{5}_{v1}$ and $X^{5}_{v2}$ from both views. We adopt a shared channel-attention mechanism, i.e., Squeeze-and-Excitation (SE)~\cite{20}, to reweight their channels in a cross-view manner and adaptively learn the inter-model relationships. Crucially, cross-multiplication is employed to process two channels instead of merely scaling each feature map by its own weights. Specifically, the attention derived from the global context of $v_1$ is used to scale the channels of $v_2$'s feature map, and vice versa.

\subsubsection{Deep Fusion: Multi-View Combination Mamba.} In the previous CVSM block, features from two views interleave to enhance each other, capturing localized cross-view interactions. However, to generate a more holistic representation that integrates critical information from both views at a deeper level, we introduce the MVCM block. Unlike CVSM, which focuses on localized feature interleaving, MVCM performs a more comprehensive fusion by aggregating and refining the outputs of the CVSM block. This transition from shallow fusion (local feature mixing) to deep fusion (global integration) ensures that the final representation effectively captures higher-level fused features from both views, as illustrated in Fig.~\ref{fig1}(iv). Specifically, the outputs $X^{5}_{v1}$ and $X^{5}_{v2}$ from the CVSM block are first fused as one of the inputs. Additionally, two separate view branches are introduced, each processing features from different views as input. Subsequently, the three branches are processed through linear layers, depth-wise convolution layers, and SiLU activation before being fed into the multi-view combination SSM. Within the multi-view combination SSM, the system matrix $\mathbf{C}_{fuse}$ information is exchanged across multiple selective scan modules to enhance multi-view feature fusion, inspired by~\cite{22,23}. The Mamba selection mechanism generates the system matrices $\mathbf{B}$, $\mathbf{C}$ and $\Delta$ from linear projection layers, facilitating the context-awareness capability of the model. As specified in Eq~\ref{eq3}, the matrix $\mathbf{C}$ decodes information from the hidden state $h_t$ to produce the output $y_t$. Therefore, we employ the matrix $\mathbf{C}_{fuse}$ of fused-branch to decode two separate view branches enhancing the model's multi-view context-awareness ability. The modified Eq.~\ref{eq3} is applicable to the multi-view combination SSM and is defined as follows:
\begin{equation}
\begingroup
\textstyle 
\label{eq5}
\begin{aligned}
y^t_{v1} &= \mathbf{C}_{fuse} h^t_{v1} + \mathbf{D}_{v1} x^t_{v1}, \quad y_{v1}=[y^1_{v1},y^2_{v1},...,y^n_{v1}],\\
y^t_{v2} &= \mathbf{C}_{fuse} h^t_{v2} + \mathbf{D}_{v2} x^t_{v2}, \quad y_{v2}=[y^1_{v2},y^2_{v2},...,y^n_{v2}],\\
y^t_{fuse} &= \mathbf{C}_{fuse} h^t_{fuse} + \mathbf{D}_{fuse} x^t_{fuse}, \quad y_{fuse}=[y^1_{fuse},y^2_{fuse},\ldots,y^n_{fuse}].
\end{aligned}
\endgroup
\end{equation}
Where, $x^t_{v1/v2/fuse}$ defines as the input at time step $t$, while $y_{v1/v2/fuse}$ represents the output of the multi-view combination SSM. After the multi-view combination SSM process, we propose a fuse-multiplication mechanism to compute channel attention on the normalized original fused view-specific features, adaptively learning the fused-view channel relationships to enhance the feature representation of a single view. Finally, additive fusion is applied to combine the three feature branches, resulting in a final holistic fused feature. 

\section{Experiments}
\subsection{Datasets and Evaluation Metrics}
\textbf{Datasets.} To evaluate the effectiveness of our approach, we conducted experiments on the MURA~\cite{13}, CheXpert~\cite{14}, and CBIS-DDSM~\cite{15} datasets. The \textbf{MURA} dataset is the musculoskeletal abnormality detection dataset, which is a binary classification task and contains two view X-ray images for each patient. We have selected patients with two views and divided the patients into random subsets for training (35,185), validation (3,158) and testing (3,283). The \textbf{CheXpert} dataset comprises chest X-ray images annotated for 13 distinct observations with positive and negative labels. We selected patients that included both frontal and lateral views and randomly divided the samples into subsets for training (23,559), validation (3,926) and testing (3,928). The \textbf{CBIS-DDSM} is a mammography dataset with craniocaudal (CC) and mediolateral-oblique (MLO) views, which is a binary classification to predict benign versus malignant cases for each CC/MLO pair. The training, validation, and test splits comprise 857, 215, and 274 CC/MLO pair samples, respectively. We applied the cropped method for CBIS-DDSM, outlined by~\cite{24}, adopting thresholding to position a fixed-size cropping window that captures the breast while minimizing the inclusion of an empty background. For all three datasets, we resized the images to 224$\times$224 prior to inputting them into the models. 

\subsection{Experimental Details}
Our XFMamba approach was implemented using PyTorch and trained on an NVIDIA A100 GPU. We adopted the Adam optimizer with the initial learning rate $1e^{-4}$ and weight decay $1e^{-5}$ to update the model parameters. The model was trained for 100 epochs with a batch size of 16. We initialized our XFMamba model with the pre-trained weights provided by VMamba~\cite{17} for the XFMamba feature encoder, constructing three variants of the model with different sizes: tiny (XFMamba-T), small (XFMamba-S), and base (XFMamba-B).

\subsection{Comparison with State-of-the-Art Methods}
To evaluate the performance, we compared the AUROC values of our XFMamba method with CVT~\cite{5}, MVC-NET~\cite{2}, MV-Swin-T~\cite{6}, MV-HFMD~\cite{25}, and BI-Mamba~\cite{26}. We follow the experimental protocol~\cite{5}, which is optimized for the cross-entropy loss and repeated over four training runs.  The quantitative results are presented in Table~\ref{tab1}, showcasing the classification performance along with a comparison of model sizes across the three datasets. Our tiny model outperforms other methods on the MURA and CheXpert datasets while maintaining the second-fewest number of parameters. Additionally, our small and base methods achieve the highest AUROC values across all three datasets. This superiority arises from the capability of our proposed XFMamba model to efficiently extract features from each individual view and while leveraging cross-view features.
\begin{table}[h!]
\centering
\caption{AUROC and model size (millions of parameters) comparisons on MURA (2 classes), CheXpert (13 classes), and CBIS-DDSM (2 classes) datasets. The mean and standard deviation of AUROC over four training runs.}
\label{tab1}
\renewcommand{\arraystretch}{1}
\footnotesize 
\begin{tabular}{cc|c|ccc}
\toprule
Method & Backbone & Params(M)   & MURA & CheXpert& CBIS-DDSM  \\ 
\hline
CVT            & Resnet18 & 39.95  & $0.876 \pm 0.003$&  $0.909 \pm 0.002$ & $0.697 \pm 0.002$ \\
BI-Mamba       & ViM-s    & \textbf{37.39}   & $0.878 \pm 0.004$ & $0.905 \pm 0.002$ & $0.611 \pm 0.003$ \\
MV-HFMD        & ViT-s    & 54.07   & $0.886 \pm 0.002$ & $0.909 \pm 0.003$ & $0.569 \pm 0.002$ \\
MVC-NET        & ResNet26 & 71.10   & $0.838 \pm 0.003$ & $0.892 \pm 0.002$ & $0.702 \pm 0.003$ \\
MV-Swin-T       & Swin-T   & 58.32   & $0.711 \pm 0.002$ & $0.882 \pm 0.005$ & $0.566 \pm 0.002$ \\
\hline
XFMamba    & VMamba-t & 39.29   & $0.898 \pm 0.002$ & $0.917 \pm 0.003$ & $0.664 \pm 0.002$ \\
XFMamba    & VMamba-s & 58.73   & $\textbf{0.910} \pm \textbf{0.003}$ & $0.918 \pm 0.002$ & $0.752 \pm 0.004$  \\
XFMamba    & VMamba-b & 90.54   & $0.904 \pm 0.004$ & $\textbf{0.919} \pm \textbf{0.002}$ & $\textbf{0.761} \pm \textbf{0.003}$\\
\bottomrule
\end{tabular}
\end{table}

\begin{figure}[htb]
    \centering
    \begin{minipage}[b]{0.45\textwidth}
        \centering
        \includegraphics[width=\textwidth]{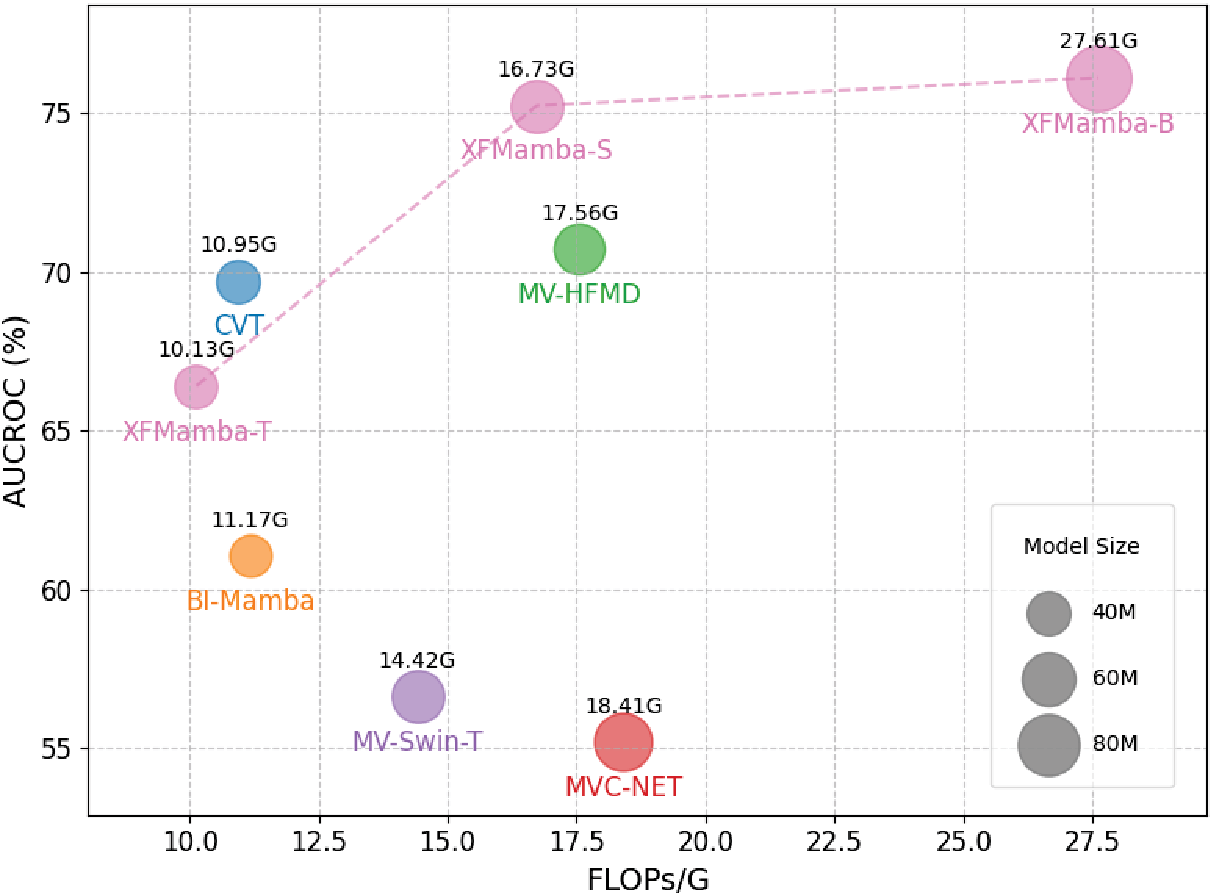}
    \end{minipage}%
    \hspace{0.015\textwidth}
    \begin{minipage}[b]{0.45\textwidth}
        \centering
        \includegraphics[width=\textwidth]{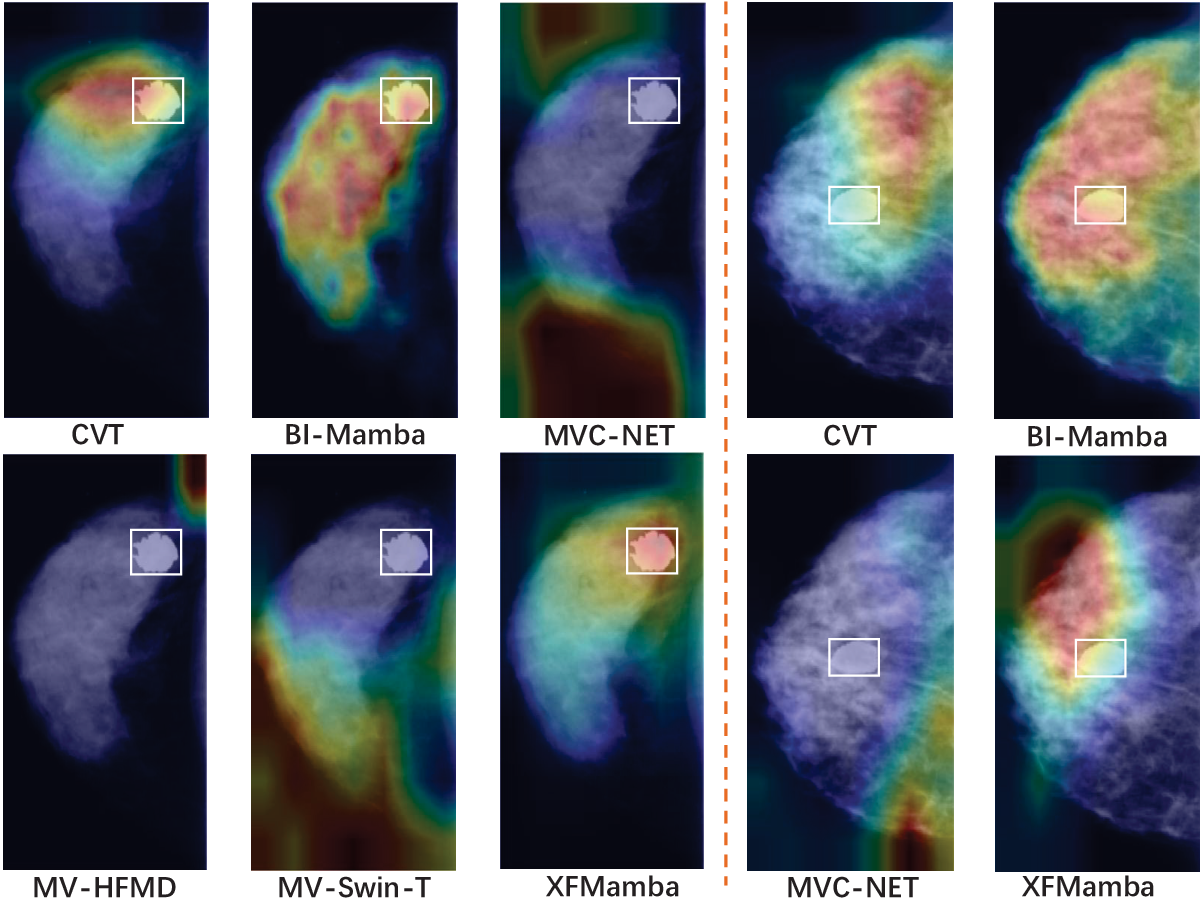}
    \end{minipage}
    \caption{ (Left) Computational complexity comparison on CBIS-DDSM dataset. The size of each circle denotes the model size, i.e., parameters. (Right) Qualitative results for successful cases and failed cases (right of the orange dotted line) using different methods on the CBIS-DDSM dataset.}
    \label{fig2}
\end{figure}

Furthermore, Fig.~\ref{fig2}(Left) compares the model computational complexity on the CBIS-DDSM dataset. The results indicate that our tiny model has the lowest FLOPs among the compared approaches while still achieving competitive performance. Fig.~\ref{fig2}(right) presents the qualitative results for GradCAM~\cite{27} heatmap examples in the CBIS-DDSM dataset against other methods. The white mask of the full mammogram image is the ROI of abnormalities. XFMamba's heatmaps are more precise and tend to focus on the ROI mask, whereas other models' heatmaps may highlight irrelevant areas. However, in failed cases, all models struggled to distinguish between abnormal and normal regions due to the limited visibility of the pathological structure.

\subsection{Ablation Study}
As illustrated in Table~\ref{tab2}, we performed an ablation study with XFMamba-S on the CheXpert dataset. 
\begin{table}[!ht]
\centering
\caption{XFMamba ablation study on CheXpert.}
\label{tab2}
\footnotesize 
\begin{tabular}{ccccccc}
\toprule
Num. & Views & Fusion Type &  CVSM-Block & MVCM-Block & AUROC$\boldsymbol{\uparrow}$\\
\midrule
1 & Lateral           & no            &  $\usym{2717}$       & $\usym{2717}$           & 0.8961       \\
2 & Frontal           & no            &  $\usym{2717}$       & $\usym{2717}$           & 0.9081 \\
3 & Frontal \& Lateral & early-fusion  &  $\usym{2717}$       & $\usym{2717}$           & 0.9090       \\
4 & Frontal \& Lateral & late-fusion   &  $\usym{2717}$       & $\usym{2717}$           & 0.9092 \\
5 & Frontal \& Lateral & cross-fusion  &  $\usym{2713}$   & $\usym{2717}$ (concat)  & 0.9129       \\
6 & Frontal \& Lateral & cross-fusion  &  $\usym{2713}$   & $\usym{2717}$ (add)     & 0.9136       \\
7 & Frontal \& Lateral & cross-fusion  &  $\usym{2717}$       & $\usym{2713}$       & 0.9144       \\
8 & Frontal \& Lateral & cross-fusion  &   $\usym{2713}$   & $\usym{2713}$       & \textbf{0.9184} \\
\bottomrule
\end{tabular}%
\end{table}
Compared to the complete model, eliminating the CSVM-Block results in a $0.40\%$ performance reduction. To evaluate the impact of removing the MVCM-Block, we test two simple fusion methods (addition and concatenation) reducing performance by $0.48\%$ and $0.55\%$, respectively. Eliminating both blocks and employing late fusion for feature integration results in a $0.92\%$ performance decline. Additionally, we compare our proposed cross-fusion model with both early-fusion and late-fusion models, finding that XFMamba outperforms them by $0.94\%$ and $0.92\%$, respectively. To further assess the effectiveness of multiple views, we compare the lateral view, frontal view, and their combination, observing that incorporating multiple views improves performance by $1.03\%$ and $2.17\%$ compared to using only the lateral or frontal view, respectively.

\section{Conclusion and Future Work}
In this paper, we have proposed a novel XFMamba network for multi-view medical image classification. Our model combines a four-stage encoder and two-stage fusion module, enabling the effective learning of single-view features and their cross-view combination. Our cross-view swapping Mamba block enriches feature across views through cross-view information integration. The multi-view combination Mamba block further enhances multi-view information fusion through a multi-view selective scan mechanism. Comparison experiments and ablation studies demonstrate the effectiveness of our approach. In future work, our XFMamba network can serve as a baseline for multi-modality models that combine different imaging modalities. We also plan to explore clinical applications where multiple views are registered, or exhibit geometric relationships, leveraging our approach to enhance feature alignment and information fusion.

\subsubsection{Acknowledgments.} This work forms part of the National Institute for Health Research Barts Biomedical Research Centre (NIHR203330).

\end{document}